\PassOptionsToPackage{hyperfootnotes=false}{hyperref}

\documentclass[11pt]{article}

\usepackage[final]{acl}

\usepackage{amsmath,amsfonts,bm}

\def\eqref#1{equation~\ref{#1}}

\def\1{\bm{1}}

\DeclareMathAlphabet{\mathsfit}{\encodingdefault}{\sfdefault}{m}{sl}
\SetMathAlphabet{\mathsfit}{bold}{\encodingdefault}{\sfdefault}{bx}{n}

\usepackage{times}
\usepackage{latexsym}
\usepackage[T1]{fontenc}
\usepackage[utf8]{inputenc}
\usepackage{microtype}
\usepackage{inconsolata}
\usepackage{graphicx}

\usepackage{booktabs}
\usepackage{multirow}
\usepackage{algpseudocode}
\usepackage{wrapfig}
\usepackage[linesnumbered,ruled,vlined]{algorithm2e}
\usepackage{multicol}
\usepackage{tablefootnote}
\usepackage{hyperref}
\usepackage{authblk}

\title{Journey Before Destination: On the importance of Visual Faithfulness in Slow Thinking}

\author[1]{\bf Rheeya Uppaal\thanks{Work done during an internship at AWS AI Labs. Correspondence to: \texttt{uppaal@cs.wisc.edu}}}
\author[2]{\bf Phu Mon Htut}
\author[2]{\bf Min Bai}
\author[2]{\bf Nikolaos Pappas} 
\author[2]{\\ \bf Zheng Qi}
\author[2]{\bf Sandesh Swamy}
\affil[1]{University of Wisconsin-Madison}
\affil[2]{AWS AI Labs}

\begin{document}
\maketitle

\begin{abstract}

Reasoning-augmented vision--language models (VLMs) generate explicit chains of thought that promise greater capability and transparency but also introduce new failure modes: models may reach correct answers via visually unfaithful intermediate steps, or reason faithfully yet fail on the final prediction. Standard evaluations that only measure final-answer accuracy cannot distinguish these behaviors.
We introduce the \emph{visual faithfulness of reasoning chains} as a distinct evaluation dimension, focusing on whether the perception steps of a reasoning chain are grounded in the image. We propose a training- and reference-free framework that decomposes chains into perception versus reasoning steps and uses off-the-shelf VLM judges for step-level faithfulness, additionally verifying this approach through a human meta-evaluation. Building on this metric, we present a lightweight self-reflection procedure that detects and locally regenerates unfaithful perception steps without any training. Across multiple reasoning-trained VLMs and perception-heavy benchmarks, our method reduces Unfaithful Perception Rate while preserving final-answer accuracy, improving the reliability of multimodal reasoning.

\end{abstract}

\section{Introduction}
\label{sec:intro}

Hallucinations in vision–language models (VLMs) are typically defined as deviations between model outputs and the underlying visual content \citep{bai2024hallucination, liu2024survey}.
While the phenomenon has been studied extensively, existing evaluations for it remain narrow. Most focus on coarse object existence in captions, 
overlooking finer elements such as counts, colors, or spatial relations that make up a large portion of visual hallucinations~\citep{gunjal2023detecting}. 
These limitations become more pronounced in reasoning based models, where intermediate steps are incorporated to solve complex tasks and provide apparent transparency into the model’s decision-making processes~\citep{li2025system}.

In text-only domains, the quality of reasoning traces has been examined in terms of their correctness, coherence, or adherence to instructions~\citep{jacovi2024chain, haollm}. 
In multimodal settings, however, these reasoning chains introduce a new axis of reliability: \emph{visual faithfulness} -- Is each step of the reasoning chain actually grounded in the image?

\begin{figure}
    \centering
    \vspace{5mm}
    \includegraphics[width=0.47\textwidth]{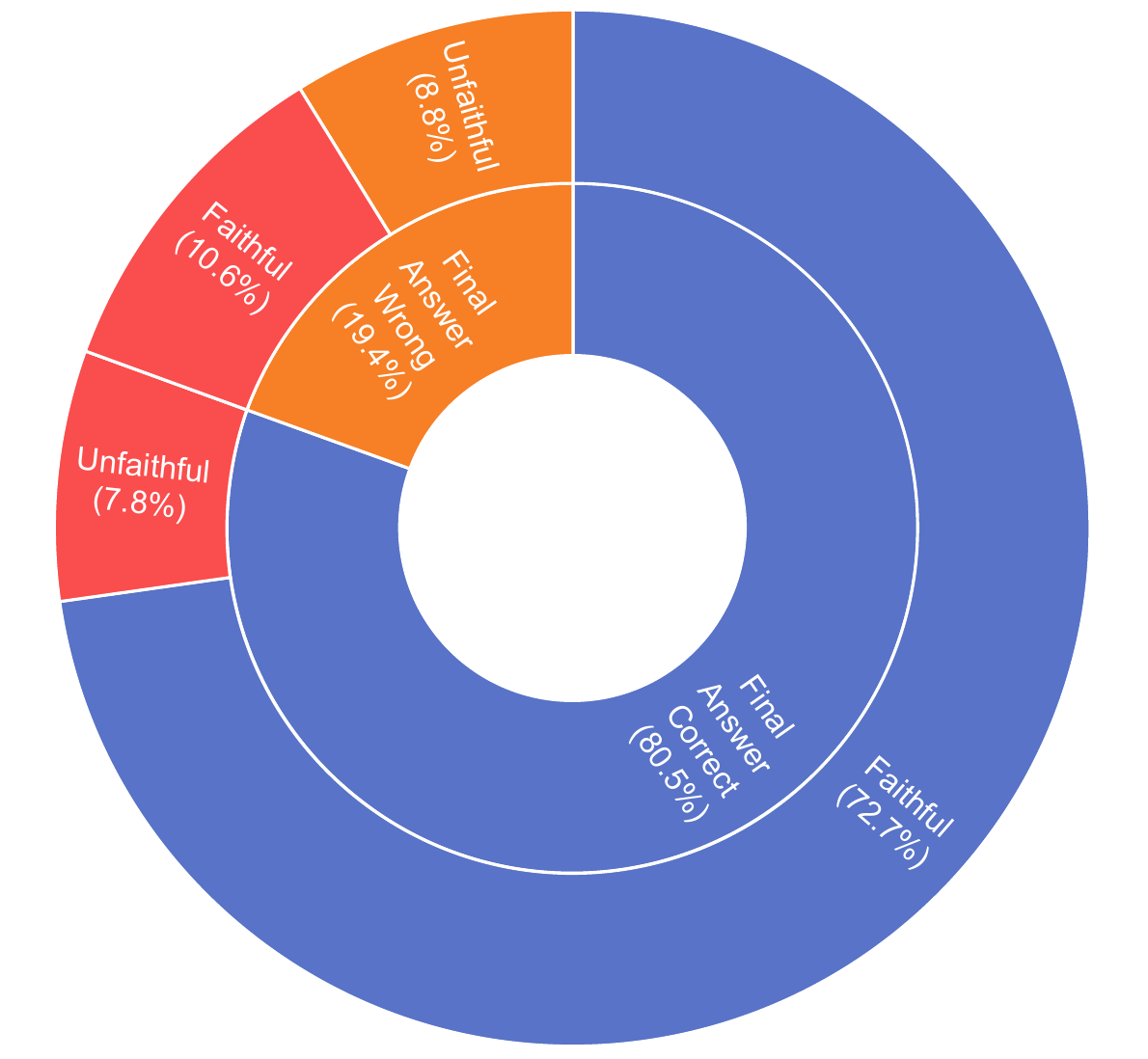}
    \caption{
    Reasoning faithfulness and final-answer accuracy diverge. 
    Correct final answers are not always grounded in the image, and incorrect answers can still reflect visually faithful reasoning. Evaluating only final accuracy therefore overlooks whether the reasoning process itself attends to the visual evidence.
    The weak correspondence between final-answer correctness and reasoning-chain faithfulness shows that accuracy metrics alone cannot capture whether a model’s reasoning genuinely reflects what it “sees.”
    }
    \label{fig:accuracy-vs-visual-faithfulness}
    \vspace{-3mm}
\end{figure}

\begin{figure*}
    \centering
    \includegraphics[width=\textwidth]{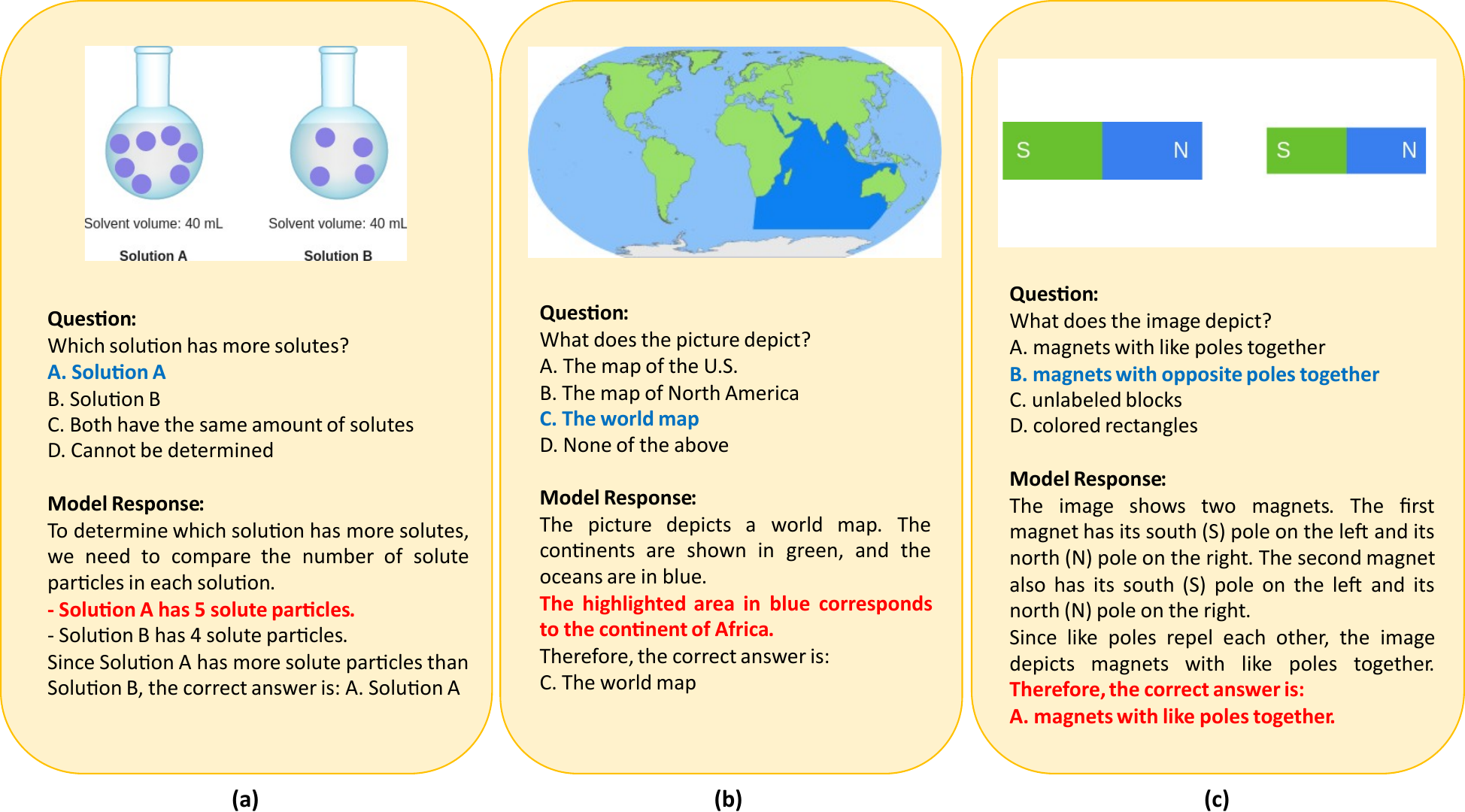} 
    \caption{  
    Reasoning-chain faithfulness does not always align with final-answer correctness.
(a–b) Visually unfaithful reasoning chains that nonetheless yield correct answers on perception tasks.
(c) A visually faithful chain producing an incorrect answer, where the error arises from reasoning rather than perception.
All responses are from the \texttt{ThinkLite-VL} model on samples from the \texttt{MMEvalPro} dataset.
    }
    \label{fig:perception-errors-in-visual-chains}
\end{figure*}

A model may produce a correct final answer while hallucinating intermediate entities, attributes, or relations that are not visually supported. Conversely, it may describe the image faithfully yet still reach an incorrect conclusion due to downstream logical mistakes. Figure~\ref{fig:perception-errors-in-visual-chains} illustrates both phenomena: visually unfaithful reasoning leading to correct predictions, and visually faithful reasoning that nonetheless yields incorrect answers. 

\begin{figure}
    \centering
    \includegraphics[width=0.4\textwidth]{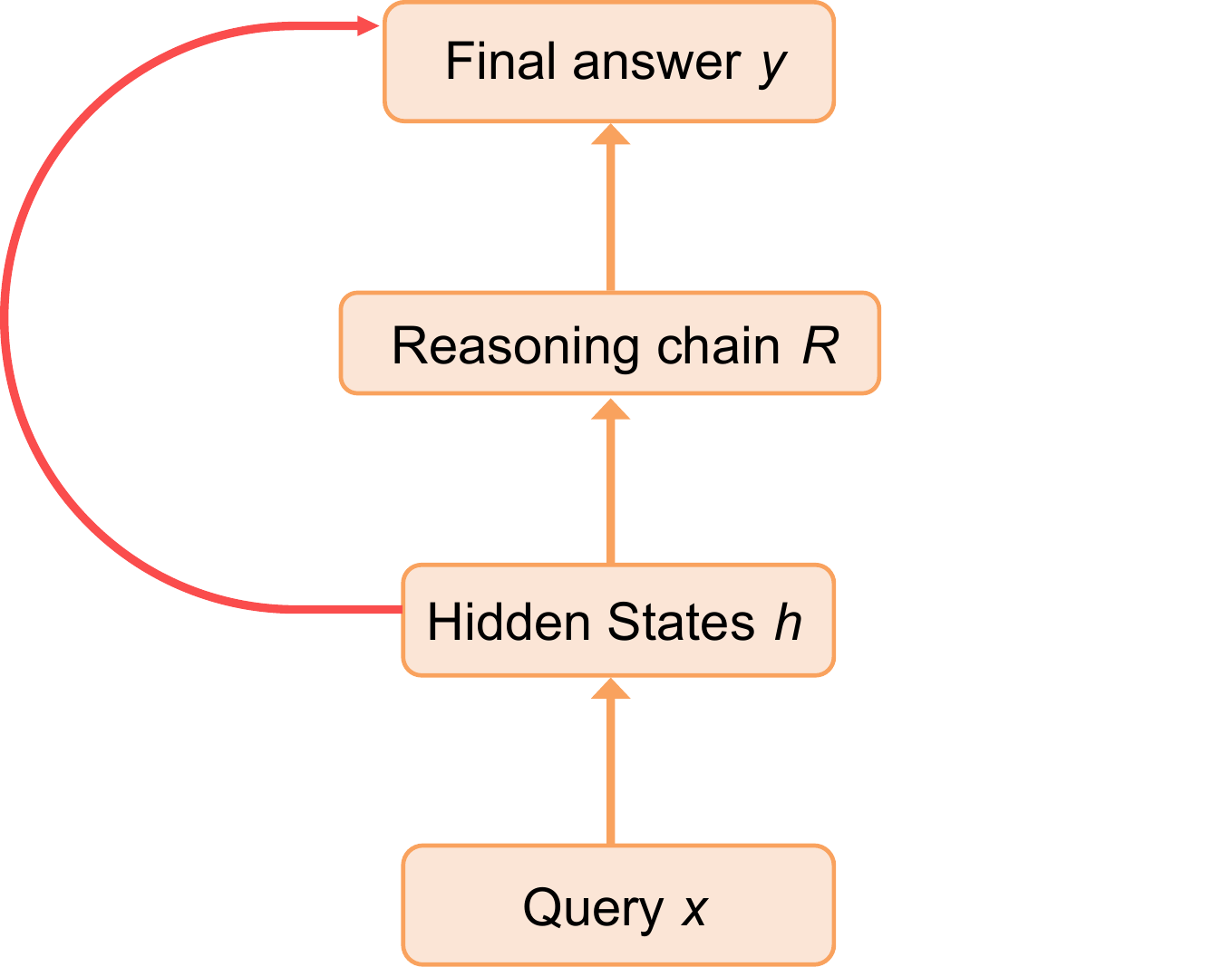} 
    \caption{Causal structure underlying final answers and reasoning traces. Many evaluation protocols assume the final answer $y$ is produced via the reasoning chain $R$ (orange arrows). 
    However, models can also map hidden features $h$ directly to $y$ (red arrow) via spurious correlations or language priors, bypassing $R$. Thus, high final-answer accuracy does not guarantee that intermediate reasoning steps are visually faithful.
    }
    \label{fig:causal-chain-of-prediction}
\end{figure}

This sheds light on a pressing issue -- existing metrics assess the hallucination rate of a VLM through its final answer accuracy on perception tasks. 
In Figure~\ref{fig:accuracy-vs-visual-faithfulness} we highlight that this measure does \textit{not} correlate with the model's hallucination rate in its reasoning chains.
Many evaluation protocols implicitly assume that the final answer $y$ is produced by following the reasoning chain $R$ (Figure~\ref{fig:causal-chain-of-prediction}). In practice, large models can shortcut this process: internal representations $h$ may map directly to the answer via spurious correlations, language priors, or pattern matching, while the chain $R$ is generated as a post-hoc justification rather than the causal basis for the decision~\citep{jiang2024interpreting, shojaee2025illusion, xia2025evaluating}. As a result, high final-answer accuracy does not guarantee that intermediate reasoning steps are visually faithful, nor that they faithfully track the model’s internal decision path. 

Existing hallucination detection approaches are simply not designed to capture hallucination rates in the complex setting of reasoning. 
These methods usually verify the existence of objects or atomic facts against the image or a ground truth list; thus treating each answer as an unordered set of facts.
By contrast, reasoning chains are a compositional trajectory -- they are long, structured, and explicitly interleave distinct perception steps (reading from the image) and reasoning steps  (operating over previously inferred facts).
Visual faithfulness is only well-defined for perception steps, yet errors in these steps can propagate through the chain and contaminate subsequent reasoning. Prior work has also shown that hallucinations become more frequent as generations grow longer and more verbose~\citep{zhai2023halle}, making it particularly important to evaluate the quality of the full reasoning trajectory rather than only its endpoint.

In this paper, we take a first step toward systematically \emph{measuring and improving the visual faithfulness of reasoning chains} in VLMs. Our contributions are:

\begin{itemize}
    \item \textit{Problem definition.} We formally highlight visual faithfulness of reasoning chains as distinct from final-answer accuracy or traditional hallucination detection, with empirical evidence that these standard metrics do not reliably capture step-level faithfulness.
    
    \item \textit{Evaluation framework.} We introduce a scalable, training- and reference-free evaluation pipeline that uses off-the-shelf VLM judges to assess visual faithfulness at the level of individual reasoning steps, and validate these metrics via a human correlation study.

    \item \textit{Mitigation method.}  We propose a lightweight self-reflection procedure that combines a when-to-intervene detector with localized regeneration of unfaithful perception steps. Our method substantially improves reasoning-chain visual faithfulness across multiple datasets and models, often with improved final-answer accuracy.

\end{itemize}

Together, our results establish reasoning-chain visual faithfulness as an essential axis for evaluating and improving reasoning-augmented VLMs, and provide concrete tools for measuring and mitigating unfaithful visual reasoning at scale.

\section{Related Work}
\label{sec:related-work}

\paragraph{Hallucinations in VLMs}
In vision–language models (VLMs), hallucinations are broadly defined as deviations between model outputs and provided visual content~\citep{bai2024hallucination, liu2024survey}. 
Such lack of visual faithfulness can arise from insufficiently diverse data during instruction tuning~\citep{li2023evaluating, wang2023amber, yu2024hallucidoctor, goyal2024context} 
or earlier training stages~\citep{wang2023amber, zhou2023analyzing, li2023evaluating, guan2024hallusionbench}; 
or limited capabilities of the vision encoder to capture fine-grained visual information~~\citep{zhang2021mme, wang2023amber, wang2024valid, liu2024mmbench, wang2024measuring}.
However, the most common cause is language dominance: VLMs tend to under-attend to the image~\citep{parcalabescu2024vision, yin2025clearsight, yangunderstanding}, 
allowing strong priors from LLM parameters to override visual signal~\citep{zhai2023halle, jiang2024interpreting, sun2024redeep, rahmanzadehgervi2024vision, liu2024reducing}.

Efforts to mitigate hallucinations have included data diversification~\citep[\textit{inter alia}]{qi2020reverie, liu2023mitigating, wang2024vigc, yu2024hallucidoctor, zhang2024reflective, zou2024look, yue2024less, hu2025prescribing}, 
activation steering to strengthen visual signal~\citep{zhai2023halle, liu2024reducing, jiang2024interpreting, yangunderstanding, yin2025clearsight, su2025activationa}, 
or other model editing techniques~\citep{jiang2024interpreting, yang2025nullu, uppaalmodel, arif2025paint}, 
and decoding-time interventions~\citep{favero2024multi, ghosh2024visual, wang2024valid, yin2024woodpecker}.
Nevertheless, most methods focus on improving visual grounding in simple discriminative or captioning settings rather than multi-step reasoning.

\paragraph{Slow Thinking VLMs}
Slow thinking is operationalized through the notion of inference-time scaling — allocating a larger token budget to allow multi-step deliberation and exploration of multiple hypotheses~\citep{li2025system}. These models demonstrate substantial gains on reasoning tasks, often trained via supervised fine-tuning (SFT) on reasoning chains~\citep{xu2024llava, zhang2024improve, deng2024enhancing, cheng2025vision, chen2024measuring} 
or reinforcement learning (RL)~\citep{wang2025unified, xiang2024atomthink, wang2025vl}.
These models consistently invoke inference-time scaling irrespective of inference time prompt structures. These training paradigms encourage self-reflection and iterative correction within the model’s generation process.

Although some recent studies attempt to enhance visual faithfulness through similar reasoning-based training~\citep{zhao2023beyond, jingfgaif, sun2024aligning, favero2024multi, zhao2025mitigating}, 
\citet{liu2025more} show that reasoning training can worsen visual faithfulness. 
Moreover, existing studies largely assess grounding only in final answers, overlooking whether intermediate reasoning steps remain visually faithful.

\paragraph{Evaluating the Quality of Reasoning Chains}
\citet{wang2023amber} and \citet{favero2024multi} show that hallucination frequency increases with longer generations, making it crucial to measure the quality of the reasoning chains, not just final answers.
In text-only settings, prior work has assessed the redundancy, relevance, and correctness of intermediate steps using verifiers~\citep{jacovi2024chain, haollm}.
\citet{shojaee2025illusion} and~\citet{xia2025evaluating} further question final answer accuracy based evaluation, showing that is does not necessarily guarantee an improvement in the
overall quality of the reasoning steps. 

However, in grounded multi-modal generation, the visual faithfulness of reasoning traces remains largely unexplored.
\citet{chen2024measuring} evaluate reasoning-step consistency but require curated atomic question sets.
In contrast, our work introduces a training and data-free framework for measuring and mitigating the visual faithfulness of reasoning chains using off-the-shelf VLM judges, establishing a scalable foundation for assessing grounded reasoning quality.

\section{Measuring the Visual Faithfulness of Reasoning Chains}
\label{sec:metric}

\subsection{Measuring Visual Faithfulness Beyond Final Answers}

Much of the existing work on visual faithfulness has focused on the limited setting of discriminative free-form answers~\citep{li2023evaluating, sun2023aligning, wang2023llm, wu2023q, zhang2021mme, liu2024mmbench, guan2024hallusionbench}. In such settings, hallucination can often be detected through simple metrics like accuracy or F1 score.
Some works extend evaluation into the generative space of image captioning tasks; however, even in this case evaluation is limited to object hallucinations, and requires a ground truth list of objects present in the image~\citep{wang2023amber, liu2023mitigating}.
More recently, limited work on reference-free evaluation in free-form answers has been proposed: \citet{liu2025more} use GPT evaluation while~\citet{jing2023faithscore} decompose an output into atomic facts and use a trained model to verify if each fact is entailed by the image.

However, notably, these approaches focus primarily on isolated facts or final answers. They do not capture whether a model’s intermediate reasoning process is visually grounded and consistent with the image. Evaluating reasoning chains poses unique challenges compared to short answers or image captions: chains are multi-step, compositional, and often combine factual grounding (or perception) with logical inference (or reasoning). A model might reach a correct final answer through unfaithful intermediate steps, or conversely, follow a faithful chain that nevertheless ends in an incorrect conclusion. 
In both cases, assessments based solely on the final output fail to reflect the true quality of visual grounding. This gap motivates our work: measuring the visual faithfulness of reasoning chains requires methods that move beyond outcome-based evaluation and instead interrogate the full reasoning trajectory.

\subsection{Measuring the Visual Faithfulness of Reasoning Chains}

Inspired by the above, we introduce a fine-grained and \emph{training and reference-free} method for evaluating the visual faithfulness of reasoning chains. 
This has two main advantages: (i) it eliminates the need for data curation for and training of task-specific classifiers or entailment models, (ii) potentially generalizing better across tasks and domains. 
Instead of building narrowly trained discriminators, we rely on the general reasoning and grounding abilities of state-of-the-art VLMs, used directly out-of-the-box.

Towards this objective, we begin by asking if state-of-the-art VLMs can serve as effective judges of visual faithfulness in reasoning chains. We propose a simple approach of using off-the-shelf VLMs as a metric and perform a meta-evaluation which demonstrates that our metric highly correlates with human judgments of faithfulness.

\paragraph{Setting} 
Given a query prompt $p$ and associated image $I$, a reasoning-trained VLM $\theta$ produces a reasoning chain $R$ and final answer $y$. The reasoning chain $R$ consists of a sequence of intermediate steps $r_1, \dots, r_t$ that alternate between referencing visual elements in $I$ (i.e. perception) and performing non-visual logical inference (i.e. reasoning). To capture this distinction, we categorize each step $r_i$ as either a \texttt{Perception} or \texttt{Reasoning} statement. Visual faithfulness is meaningful only for \texttt{Perception} steps, since these directly claim to ground information in the image. Accordingly, we define each perception step as either \texttt{Faithful} if it accurately reflects the visual content, or \texttt{Unfaithful} if it introduces hallucinated or incorrect visual details. This step-level distinction provides the foundation for our evaluation method, which requires a judge to disentangle perception from reasoning and assess the grounding of each perception step.

\begin{algorithm}[t]

\caption{Evaluation of Reasoning Chain Visual Faithfulness through a Judge }
\label{alg:faithfulness-judge}
\KwIn{Prompt $p$, Image $I$, VLM $\theta$, Judge $J$}
\KwOut{Annotated sequence $\{(r_t, \texttt{type}_t, \texttt{faith}_t)\}_{t=1}^T$}

Get VLM generation: $(R, y) \gets \theta(p, I)$\;
Segment reasoning chain $R$ into steps: $\{r_1, r_2, \ldots, r_T\} \gets \texttt{Segment}(R)$\;
Judge precomputes visual context: $\hat{I}_J \gets \texttt{GroundImage}(I)$\;

\For{$t \gets 1$ \KwTo $T$}{
    \eIf{$r_t$ references visual content in $I$}{
        $\texttt{type}_t \gets \textsc{Perception}$\;
        \eIf{$s_t$ is grounded in $I$}{
            $\texttt{faith}_t \gets \textsc{Faithful}$\;
        }{
            $\texttt{faith}_t \gets \textsc{Unfaithful}$\;
        }
    }{
        $\texttt{type}_t \gets \textsc{Reasoning}$\;
        $\texttt{faith}_t \gets \textsc{N/A}$\;
    }
}
\Return{$\{(r_t \texttt{type}_t, \texttt{faith}_t)\}_{t=1}^T$}
\end{algorithm}

\paragraph{Method} 
We leverage state-of-the-art VLM judges to perform fine-grained evaluations of reasoning chains at the step level. Given a judge model $J$ and a reasoning chain $R = \{r_1, \dots, r_t\}$ generated by a VLM  $\theta$, the judge is tasked with segmenting the chain into individual steps and assigning two labels to each $r_i$: a type label (\texttt{Perception} or \texttt{Reasoning}) and, when applicable, a faithfulness label (\texttt{Faithful} or \texttt{Unfaithful}). Since only perception steps are expected to reference the image, faithfulness is evaluated exclusively for these cases. To enhance reliability, we first ground the judge in the visual content by prompting it to produce a detailed description of the image $I$ prior to annotation. This auxiliary description is used internally by the judge to anchor subsequent assessments, ensuring that each step is evaluated with respect to the actual visual evidence rather than generic priors. The full evaluation procedure is summarized in Algorithm~\ref{alg:faithfulness-judge}.

\begin{table}
  \centering
   \resizebox{0.51\textwidth}{!}{
  \begin{tabular}{lcc}
    \toprule
    \multirow{2}{*}{\textbf{Judge Model}} & \multicolumn{2}{c}{\textbf{Correlation}} \\ 
    & \textbf{Perception} & \textbf{Faithfulness} \\
    \midrule
    \texttt{LLaVA-NeXT} & 0.54 & 0.45\\ 
    \texttt{Qwen2.5-VL-72B-Instruct} & 0.94 & 0.66  \\
    \texttt{Claude 3.7 Sonnet} & 0.87 & 0.66 \\ 
    \texttt{Claude 4 Sonnet} & 0.93 & \textbf{0.69} \\ 
    \bottomrule
  \end{tabular}
  }
  \caption{Comparison of various Judge models on the task of measuring visual faithfulness. The labels of each judge are compared against two sets of human annotations, using ICC 3-1 as a correlation measure. Correlations above 0.6 are considered acceptable.}
  \label{tab:judge-comparisons}
\end{table}

\begin{table}
  \centering
   \resizebox{0.5\textwidth}{!}{
  \begin{tabular}{lcc}
    \toprule
    \multirow{2}{*}{\textbf{Configuration}} & \multicolumn{2}{c}{\textbf{Correlation}} \\ 
    & \textbf{Perception} & \textbf{Faithfulness} \\
    \midrule
    Vanilla & 0.93 & 0.66 \\ 
    + Grounding & 0.93 & \textbf{0.69}  \\ 
    + Grounding + Rationales & 0.91 & 0.65\\ 
    \bottomrule
  \end{tabular}
  }
  \caption{Using the best judge model of \texttt{Claude 4 Sonnet}, we measure its correlation with human judgment against various prompting styles.}
  \label{tab:best-judge-configuration}
\end{table}

\paragraph{How well calibrated are VLM judges?}
We evaluate the calibration of several widely used VLM judges~\citep{lmarena2024}, including \texttt{LLaVA-NeXT}~\citep{liu2024llavanext}, \texttt{Qwen2.5-VL-Instruct}~\citep{bai2025qwen2}, and \texttt{Claude Sonnet 3.7} and \texttt{4}~\citep{anthropic2024claude}. 
To assess their reliability, we measure the extent to which each model’s ratings of reasoning chain faithfulness align with human annotations. 
Specifically, we collect 300 random samples from the \texttt{MMEval-Pro} benchmark~\citep{huang2024mmevalpro}, spanning math, science, and natural image domains. For each sample, VLM generations (acquired from a 7B reasoning model) are annotated both automatically (by the VLM judges) and manually (by two human annotators). We then compute the Intraclass Correlation Coefficient (specifically, ICC(3,1))~\citep{koch2004intraclass} between model and human ratings. We adopt ICC rather than agreement measures such as Cohen’s~\citep{cohen1960coefficient} or Fleiss’ Kappa~\citep{fleiss1971measuring}, since ICC is more appropriate for continuous judgments: it accounts not only for agreement in categorical assignment but also for the magnitude of differences in ratings~\citep{klie2024analyzing}. As shown in Table~\ref{tab:judge-comparisons}, \texttt{Claude Sonnet 4} achieves the highest correlation with human judgments, indicating its superior calibration as a faithfulness judge among the models evaluated.
More details can be found in Appendix~\ref{sec:appendix-vlm-judge-details}.

We further investigate how 
the performance of \texttt{Claude Sonnet 4}
varies under different prompting configurations. Table~\ref{tab:best-judge-configuration} reports the ICC values obtained across some basic prompting variants (described in Appendix~\ref{sec:appendix-vlm-judge-details}). The results suggest that increasingly complex prompts (Grounding + Rationales) may worsen correlation, and we thus use the Grounding prompt as part of our final evaluation method.

\section{Improving the Visual Faithfulness of Reasoning Chains}
\label{sec:mitigation}

\paragraph{A \textit{When} and \textit{How} Problem}
Improving visual faithfulness in reasoning chains requires addressing two distinct questions: \emph{when} should an intervention occur, and \emph{how} should the model be guided once an unfaithful step is detected? We explicitly separate these questions because reasoning chains are typically long and alternate between \texttt{Perception} and \texttt{Reasoning} steps. A global intervention strategy that applies corrections indiscriminately risks disrupting reasoning ability, while overly narrow strategies may fail to catch unfaithful references. Instead, interventions should be targeted only at \texttt{Perception} steps that are identified as unfaithful, thereby minimizing collateral effects on downstream reasoning. In addition, we deliberately focus on \textit{training-free} mitigation strategies. Such methods are modular, lightweight, and easily applicable across different models and tasks without the need for task-specific fine-tuning.

\paragraph{Self-Reflection as a Mitigation Strategy}
Our proposed approach is based on self-reflection, motivated by the observation that models often exhibit higher variance in their outputs when hallucinating~\cite{farquhar2024detecting}. The method operates in two stages. First, a detector function monitors each reasoning step in the chain and flags it as unfaithful when it fails to align with the visual evidence (\textbf{\textit{when}} to intervene). Second, once an unfaithful step is detected, the model is prompted to regenerate that portion of the chain with explicit instructions to ground its description in the image (\textbf{\textit{how}} to intervene). For example, if a model incorrectly claims that ``a dog is present in the image'' when no dog exists, the detector identifies this as unfaithful and triggers a regeneration step. The model is then instructed to re-describe the scene, producing a corrected perception such as ``no animals are present in the image.'' This localized regeneration preserves the integrity of faithful steps while correcting errors where they occur. The complete procedure is formalized in Algorithm~\ref{alg:self-reflection}.

Given an input prompt $p$ and image $I$, the VLM $\theta$ generates its answer $(R, y)$. The detector function $D$ returns the index of the first unfaithful step $i$ (or $-1$ if no steps are unfaithful). Following this, the VLM is prompted with $(p, I, r_1 \dots r_i, p_r)$ to regenerate a faithful $r_i$, where $p_r$ specifies the unfaithfulness of $r_i$. The regeneration process is repeated until $r_i$ is faithful or a retry limit $K$ is reached.\footnote{The retry limit prevents unbounded regeneration and avoids unnecessary inference cost when a step remains unfaithful after multiple regenerations. As shown in Figure~\ref{fig:self-reflection-latency}, ~90\% of successful corrections occur within three retries, so additional attempts yield negligible benefit.}
After this, the corrected and partial reasoning chain $r_1 \dots r_i'$ is fed to the VLM $\theta$ to generate the remaining reasoning chain $r_{i+1} \dots r_{t}$, restarting the self-reflection process. This continues until the last reasoning step $r_T$ is checked by $D$.

\begin{algorithm}[t]
\caption{Self-Reflection with Reasoning Trained VLMs} 
\label{alg:self-reflection}
\KwIn{Prompt $p$, Image $I$, VLM $\theta$, Detector $D$, Regeneration Prompt $p_r$, Retry Limit $K$}
\KwOut{Faithful Reasoning Chain $\tilde{R}$}

$\tilde{R} \gets \emptyset$\;
$i \gets 0$

\Repeat{reasoning complete}{
    Generate reasoning segment: $(r_{i+1}, \ldots, r_t) \gets \theta(p, I, \tilde{R})$\;
    Detect first unfaithful step: $j \gets D(r_{i+1}, \ldots, r_t)$\;

    \eIf{$j = -1$}{
        $\tilde{R} \gets \tilde{R} \cup (r_{i+1}, \ldots, r_t)$\;
    }{
        Regenerate $r_{i+j}$ with retry limit: \\
        \For{$k \gets 1$ \KwTo $K$}{
            $r_{i+j}' \gets \theta(p, I, r_1, \dots, r_{i+j} \mid p_r)$ \\
            \If{$D(r_{i+j}') = -1$}{
                break\;
            }
        }
        $\tilde{R} \gets \tilde{R} \cup (r_{i+1}, \ldots, r_{i+j}')$\;
    }
}
\Return{$\tilde{R}$}
\end{algorithm}

\section{Experimental Setup}
\label{sec:experimental-setup}

\paragraph{Models}
To ensure consistent generation of reasoning traces during inference, we use the  
following 7B reasoning trained models: 
\texttt{ThinkLite-VL}~\citep{wang2025sota}, \texttt{OpenVLThinker}~\cite{deng2025openvlthinker}, 
\texttt{MM-Eureka}~\cite{meng2025mm} and 
\texttt{Ocean-R1}~\cite{ming2025ocean}. 
All models have been trained from \texttt{Qwen2.5-VL-7B-Instruct}~\citep{team2024qwen2},
and have been selected to provide a uniform sampling over methods used for reasoning training, as well as domains the training was performed on.
More details on these models can be found in Appendix~\ref{sec:app-models-datasets}.

\paragraph{Datasets}
We use three popular perception benchmarks in our study: 
The perception split of \texttt{MMEvalPro}~\citep{huang2024mmevalpro}, \texttt{MMVP}~\citep{tong2024eyes} and \texttt{HallusionBench}~\citep{guan2024hallusionbench}.
All datasets are framed as multiple choice questions, with ground truth final answers provided. 
More details can be found in Appendix~\ref{sec:app-models-datasets}.

\paragraph{Measuring Model Performance}
We measure two facets of the VLM's generated answer: 
\begin{itemize}
    \item \textbf{Visual faithfulness per reasoning chain sentence: }
    Following Section~\ref{sec:metric}, we use \texttt{Claude 4 Sonnet} as a judge to evaluate the faithfulness of each sentence (or step) in the reasoning chain. We report Unfaithful Perception Rate (UPR), which is simply the fraction of unfaithful perception steps. In other words, 
    \begin{align*}
        UPR = \frac{\text{Number of Unfaithful Sentences}}{\text{Number of Perception Sentences}}
    \end{align*}
    These figures are calculated on a dataset level. A higher UPR indicates a higher rate of unfaithfulness in the visual reasoning, while a lower UPR suggests better alignment between perception sentences and the image content.

    \item \textbf{Final Answer Accuracy:}
    To ensure our method does not degrade original capabilities of the model, we measure the correctness of the final answer provided by the model after the reasoning chain. For this, the model's selected option on the MCQ question is compared against the Ground Truth option.
\end{itemize}

\section{Identifying When to Intervene}
\label{sec:when-to-intervene}

To determine when during generation to apply an intervention, we compare several hallucination‐detection strategies.

\paragraph{Detection Strategies}
White-box approaches use internal signals such as attention~\citep{zhang2024dhcp,huang2024opera}, logits, or hidden states~\citep{jiang2024interpreting}.
They are training-free but offer coarse, unstable signals due to model complexity~\citep{chen2025survey}.
Black-box methods rely on surface behavior, including similarity matching, uncertainty~\citep{zhang2024vl},
or trained auxiliary model judgments~\citep{jing2023faithscore,nguyen2025cutpaste,liu2023mitigating,liu2024mmbench,kaul2024throne,wang2023evaluation}.
They generalize better and are more reliable across tasks.

\paragraph{Experimental Setup}
We sample 1200 examples from the perception split of \texttt{MMEval-Pro}, holding out 500 for detector tuning.
Training focuses on early reasoning steps due to the long-tailed distribution of chain lengths (Figure~\ref{fig:long-tail-when}).
Experiments use \texttt{ThinkLite-VL-7B}, and F1 is computed on the \texttt{Unfaithful} class using \texttt{Claude 4 Sonnet}–derived gold labels. More details can be found in Appendix~\ref{sec:appendix-detection-methods}.

\paragraph{Results}

White-box methods perform poorly, reflecting weak internal calibration in 7B models.
Training-based detectors overfit early steps and degrade over time due to concept drift (Appendix~\ref{sec:appendix-detection-methods}), limiting generality.
Auxiliary-model detectors remain robust and achieve the highest F1, so we adopt this approach as our \emph{when} detector.
Results are shown in Table~\ref{tab:when-methods}.

\begin{table}
  \centering
  \resizebox{0.49\textwidth}{!}{
  \begin{tabular}{ccc}
    \toprule
    \multirow{2}{*}{\textbf{Method}} & \multicolumn{2}{c}{\textbf{F1 Score ($\uparrow$)}} \\ 
    & \textbf{Faithful Class} & \textbf{Unfaithful Class}\\
    \midrule
    SAPLMA & 74.9 & 25.4 \\
    HaloScope & 91.5 & 14.9 \\
    kNN & 67.5 & 8.9 \\
    \midrule
    Prompting & 84.8 & 30.8 \\
    Auxiliary Model & \textbf{98.6} & \textbf{97.8} \\
    \bottomrule
  \end{tabular}
  }
  \caption{Comparison of \emph{when} to intervene methods using the \texttt{ThinkLite-VL} (7B) model. Results show that hallucination detection remains challenging for a 7B VLM given limited and imbalanced training data, while a stronger auxiliary model (\texttt{Claude 3.7}) achieves substantially better performance.}
  \label{tab:when-methods}
\end{table}

\begin{figure*}[ht]
    \centering
    \includegraphics[width=0.4\textwidth]{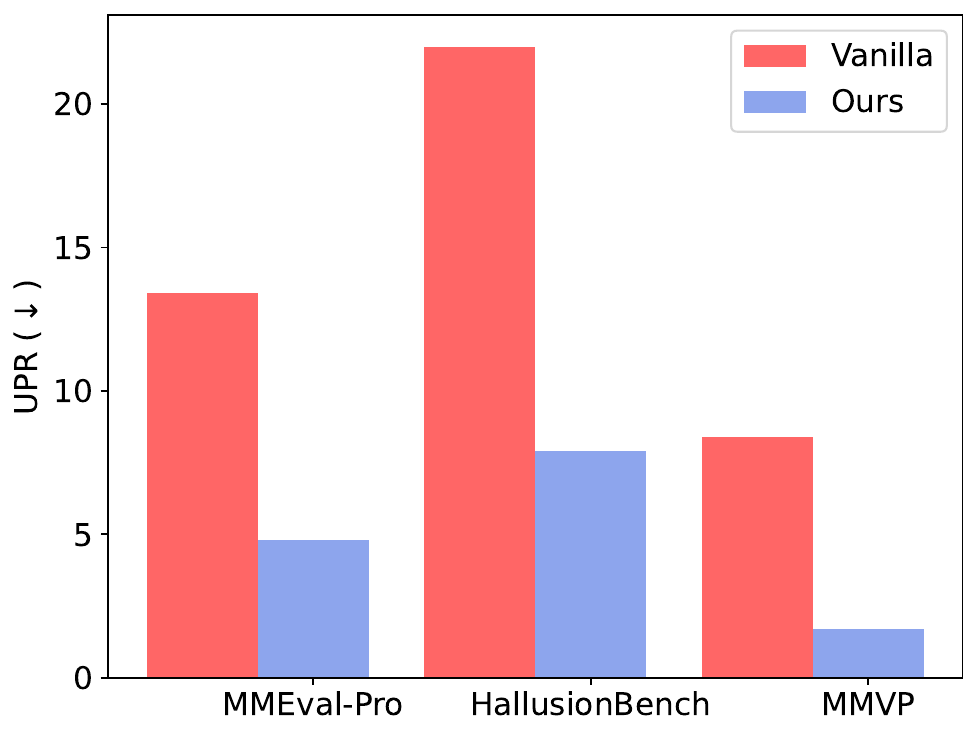}
    \includegraphics[width=0.4\textwidth]{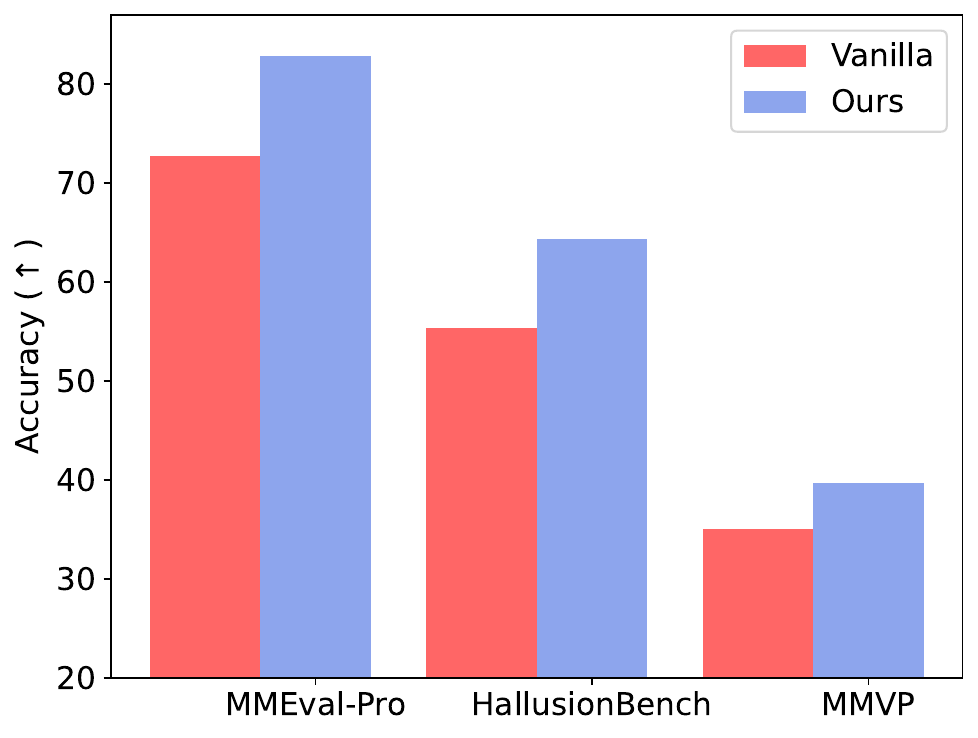}
    \caption{Impact of our method on the visual faithfulness of reasoning chains.
    Both methods (vanilla and ours) use the same underlying model (\texttt{ThinkLite VL}).
    Our method significantly reduces UPR, while also improving final answer accuracy.
    All numbers are reported as percentages.}
    \label{fig:main-result}
\end{figure*}

\section{Self-Reflection Improves Visual Faithfulness in Reasoning Chains}
\label{sec:results}

\paragraph{Self-Reflection Enhances Faithfulness and Accuracy}
Figure~\ref{fig:main-result} 
shows that our self-reflection strategy substantially improves the visual faithfulness of reasoning chains across datasets.
Interestingly, final-answer accuracy also rises, suggesting that grounding intermediate reasoning steps strengthens overall task performance.
In Appendix~\ref{sec:appendix-self-reflection-details}, we see this trend holds for various reasoning-trained models.

\begin{figure}[ht]
    \centering
    \includegraphics[width=0.4\textwidth]{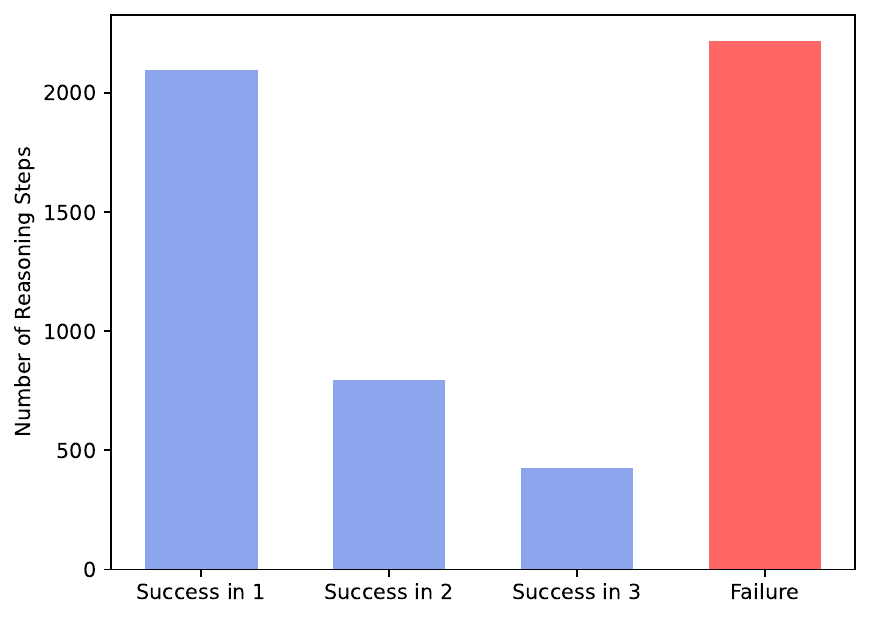}
    \caption{Breakdown of self-reflection outcomes by number of regeneration attempts.
    Most unfaithful steps are corrected within one regeneration, and over 90\% within three.
    The remaining unresolved cases correspond to instances where the model likely lacks the visual knowledge to generate a faithful description, indicating that the reflection process reaches its natural limit rather than incurring inefficiency.}
    \label{fig:self-reflection-latency}
\end{figure}

\paragraph{Importance of Knowing \textit{When} to Intervene}
The detector function in Algorithm~\ref{alg:self-reflection} is key to effective reflection.
As shown in Table~\ref{tab:ablation-on-when-in-mitigation}, 
replacing the detector function from \texttt{Claude 3.7} to the weaker \texttt{Qwen2.5-VL-72B-Instruct} shows a drop in accuracy on the \texttt{MMVP} perception task; 
and replacing our detector function to with a simple self-assessment prompt causes a sharp drop in unfaithful‐step recovery, 
confirming that precise intervention timing is critical.

\paragraph{Latency of the Self-Reflection Method}
Although self-reflection adds additional forward passes, it remains relatively efficient in practice.
As shown in Figure~\ref{fig:self-reflection-latency}, most fixable unfaithful steps are corrected within a single regeneration, and over 90\% of successful corrections occur within three attempts.
The remaining unresolved cases correspond to instances where the model likely lacks the
visual knowledge to generate a faithful description, indicating that the reflection process reaches its natural limit rather than incurring inefficiency.

\begin{table}
  \centering
  \resizebox{0.49\textwidth}{!}{
  \begin{tabular}{cccc}
    \toprule
    \textbf{Detector Function} & \textbf{UPR ($\downarrow$)} & \textbf{Acc ($\uparrow$)} \\
    \midrule
    None (Vanilla) & 8.4 & 35.0\\
    \texttt{Claude 3.7 Sonnet} & \textbf{1.7} & \textbf{39.7} \\ 
    \texttt{Qwen2.5-VL-72B-Instruct} & 2.1 & 32.0 \\
    \texttt{ThinkLite-VL 7B} & 6.5 & 33.3 \\
    \bottomrule
  \end{tabular}
  }
  \caption{Impact of the detector function $D$ (i.e. \textbf{\textit{when}} to intervene module) on the reasoning chain visual faithful faithfulness and final answer accuracy, on the \texttt{MMVP} perception task. The \texttt{ThinkLite-VL} model is used for generation.
  All numbers are recorded as percentages.
  }
  \label{tab:ablation-on-when-in-mitigation}
\end{table}

\section{Discussion}
\label{sec:conclusion}

This work identifies visual faithfulness of reasoning chains as a distinct dimension of performance for reasoning-oriented VLMs. While prior evaluations primarily emphasize final-answer accuracy, we show that such metrics do not determine whether intermediate reasoning steps are actually grounded in the image.

To address this gap, we introduce both a simple evaluation metric and a lightweight self-reflection procedure for improving step-level faithfulness. Across models and datasets, this approach consistently strengthens visual grounding and, in many settings, also improves final-task accuracy. These findings suggest that encouraging faithful reasoning can enhance not only model transparency, but also reliability on downstream tasks.

At the same time, this work should be viewed as an initial step. The proposed method is intentionally lightweight—training-free and broadly applicable across model families—but it also has clear limitations. By formalizing the problem, introducing an evaluation metric, and demonstrating a simple yet effective mitigation strategy, we aim to establish a foundation for subsequent work. In this sense, the framework serves as a stepping stone toward richer supervision signals, improved training paradigms, and ultimately models whose reasoning is not only accurate, but also transparent and visually grounded.

\section*{Limitations}
Our framework is simple and training-free, but several limitations remain.

\paragraph{Inference efficiency}
Self-reflection introduces extra forward passes, adding latency relative to a single-pass baseline. Yet this overhead is bounded (capped at three regenerations) and far lighter than retraining or collecting new data. Most correctable errors are resolved within one regeneration, making the method efficient in practice. Optimizations such as KV-cache reuse, partial decoding, or adaptive stopping could further reduce runtime.

\paragraph{Dependence on auxiliary models}
Our detection approach relies on a strong external VLM. While effective, this may limit accessibility. While we already show that strong open-source models perform comparably to closed-source ones, exploring lighter detectors or self-checking mechanisms would make the approach more widely usable.

\paragraph{Scope of evaluation}
We study perception-heavy reasoning tasks; generalizing to broader settings such as planning or multimodal dialogue is deferred to future work.

\bibliography{custom}
\clearpage
\appendix

\section{Ethical Considerations}
\label{sec:ethical-considerations}

Our primary objective is to enhance the safe utility of Large Language Models (LLMs) by reducing the potential harm caused by their outputs. By prioritizing the development of mechanisms to curtail hallucinations, we aim to contribute to a more responsible and ethical deployment of LLMs in various applications, thereby safeguarding against the propagation of misinformation and promoting the creation of safer digital environments.

Our study does not involve any human subjects or violation of legal compliance. We do not anticipate any potentially harmful consequences to our work. All of our experiments are  conducted using publicly available datasets. Our code shall be released for reproducibility. Through our study and releasing our code, we hope to raise stronger research and societal awareness towards building safe and robust language models.
 
\section{Models and Datasets}
\label{sec:app-models-datasets}

\paragraph{Reasoning VLMs}
We use the models 
\texttt{ThinkLite-VL}\footnote{\url{https://huggingface.co/russwang/ThinkLite-VL-7B}}~\cite{wang2025sota}, 
\texttt{OpenVLThinker}\footnote{\url{https://huggingface.co/ydeng9/OpenVLThinker-7B}}~\cite{deng2025openvlthinker}, 
\texttt{MM-Eureka}\footnote{\url{https://huggingface.co/FanqingM/MM-Eureka-Qwen-7B}}~\cite{meng2025mm} and
\texttt{Ocean-R1}\footnote{\url{https://github.com/VLM-RL/Ocean-R1}}~\citep{ming2025ocean}.
All models have been trained from \texttt{Qwen2.5-VL-7B-Instruct}~\citep{team2024qwen2}.
More details on these models can be found in Table~\ref{tab:model-details}.

\paragraph{Judge Models}
We shortlist commonly used vision-language judge models in our study: 
\texttt{LLaVA-NeXT}\footnote{\url{https://huggingface.co/llava-hf/llava-v1.6-34b-hf}}~\citep{liu2024llavanext}, 
\texttt{Qwen2.5-VL-72B-Instruct}\footnote{\url{https://huggingface.co/Qwen/Qwen2.5-VL-72B-Instruct}}~\citep{bai2025qwen2},
\texttt{Claude 3.7 Sonnet}\footnote{\url{https://www.anthropic.com/news/claude-3-7-sonnet}}~\citep{anthropic2024claude},
\texttt{Claude 4 Sonnet}\footnote{\url{https://www.anthropic.com/claude/sonnet}}~\citep{anthropic2024claude}.
More details can be found in Table~\ref{tab:judge-model-details}.

\paragraph{Datasets}
We use three popular perception benchmarks in our study: 
The perception split of \texttt{MMEvalPro}\footnote{\url{https://huggingface.co/datasets/MM-Diagnose/MMEvalPro}}~\citep{huang2024mmevalpro},
\texttt{MMVP}\footnote{\url{https://huggingface.co/datasets/MMVP/MMVP}}~\citep{tong2024eyes} and 
\texttt{HallusionBench}\footnote{\url{https://huggingface.co/datasets/rayguan/HallusionBench}}~\citep{guan2024hallusionbench}.
More statistics about each dataset is available in Table~\ref{tab:dataset-details}. The listed datasets are intended for research purposes only. We do not make any commercial use of them.

\paragraph{Implementation Details} All experiments were run on A100 GPUs. We use HuggingFace for all our implementations, and will publically release our code.

\begin{table*}[ht]
\centering
    \begin{tabular}{lcccc}
    \toprule
        \textbf{Model} & \textbf{Size} & \textbf{Training Domain} & \textbf{Training Method} & \textbf{License} \\ 
        \midrule
        \texttt{ThinkLite-VL} & 7B & Math & RL (GRPO) & Unknown\\
        \texttt{OpenVLThinker} & 7B & Math & SFT + RL (GRPO) &  Apache 2.0\\
        \texttt{MM-Eureka} & 7B, 32B & Math & RL & Apache-2.0 \\ 
        \texttt{Ocean-R1} & 7B & General & RL & Unknown \\
        \bottomrule
    \end{tabular}
\caption{Reasoning Trained Vision Language Models used in our study. All models are accessed through HuggingFace\tablefootnote{\url{https://huggingface.co/models}}.}
\label{tab:model-details}
\end{table*}

\begin{table*}[ht]
\centering
    \begin{tabular}{lll}
    \toprule
        \textbf{Model} & \textbf{Access} & \textbf{License} \\ 
        \midrule
        \texttt{LLaVA-NeXT} & HuggingFace & Llama 2 Community License Agreement\\
        \texttt{Qwen2.5-VL-72B-Instruct} & HuggingFace & Apache 2.0\\
        \texttt{Claude 3.7 Sonnet} & API & Proprietary  \\ 
        \texttt{Claude 4 Sonnet} & API & Proprietary  \\
        \bottomrule
    \end{tabular}
\caption{Judge Models used in our study.}
\label{tab:judge-model-details}
\end{table*}

\begin{table*}[ht]
\centering
    \begin{tabular}{llll}
    \toprule
        \textbf{Dataset} & \textbf{Language} & \textbf{License} & \textbf{Number of Samples} \\ 
        \midrule
        \texttt{MMEvalPro} (Perception Split) & English & CC BY-SA 4.0 & 2200 \\
        \texttt{MMVP} & English & MIT & 300 \\
        \texttt{HallusionBench} & English & BSD 3-Clause & 1000 \\
        \bottomrule
    \end{tabular}
\caption{Artifacts used in our study. The dataset statistics report the values used in our study.}
\label{tab:dataset-details}
\end{table*}

\begin{table*}
  \centering
  \begin{tabular}{llcc}
    \toprule
    \multirow{2}{*}{\textbf{Dataset}} & \multirow{2}{*}{\textbf{Judge Model}} & \multicolumn{2}{c}{\textbf{Correlation}} \\ 
    & & \textbf{Perception} & \textbf{Faithfulness} \\
    \midrule
    \multirow{4}{*}{MMMU} & \texttt{LLaVA-NeXT} & 0.48 & 0.41 \\
                          & \texttt{Qwen2.5-VL-72B-Instruct} & 0.92 & 0.82 \\
                          & \texttt{Claude 3.7 Sonnet} & \textbf{0.95} & 0.73 \\
                          & \texttt{Claude 4 Sonnet} & 0.92 & \textbf{0.82}\\ 
    \midrule
    \multirow{4}{*}{ScienceQA} & \texttt{LLaVA-NeXT} & 0.52 & 0.45 \\
                          & \texttt{Qwen2.5-VL-72B-Instruct} & \textbf{0.9} & 0.59\\
                          & \texttt{Claude 3.7 Sonnet} & 0.82 & \textbf{0.67} \\
                          & \texttt{Claude 4 Sonnet} & 0.8 & 0.56 \\ 
    \midrule
    \multirow{4}{*}{MathVista} & \texttt{LLaVA-NeXT} & 0.56 & 0.46 \\
                          & \texttt{Qwen2.5-VL-72B-Instruct} & \textbf{0.96} & 0.62 \\
                          & \texttt{Claude 3.7 Sonnet} & 0.86 & 0.63 \\
                          & \texttt{Claude 4 Sonnet} & \textbf{0.96} & \textbf{0.73}\\ 
    \bottomrule
  \end{tabular}
  \caption{Comparison of various Judge models on the task of measuring visual faithfulness. The labels of each judge are compared against two sets of human annotations, using ICC 3-1 as a correlation measure. Correlations above 0.6 are considered acceptable as per~\citet{koo2016guideline}.}
  \label{tab:judge-comparisons-full}
\end{table*}

\section{Measuring Visual Faithfulness with VLM Judge Models}
\label{sec:appendix-vlm-judge-details}

In this section, we include supplementary information to Section~\ref{sec:metric}. 

\paragraph{Evaluation Data}
The data used for the human correlation study is sampled from the \texttt{MMEval-Pro} dataset~\citep{huang2024mmevalpro}, which consists of three splits sources from existing VLM benchmarks:
\texttt{MMMU}~\citep{yue2024mmmu}, \texttt{ScienceQA}~\citep{lu2022learn} and \texttt{MathVista}~\citep{lumathvista}. 
Similar to~\citet{jing2023faithscore}, we choose 100 samples at random from each split, following which the corresponding model responses are generated using the \texttt{ThinkLite-VL} (7B) model. Specifically, given a prompt image pair $(p, I)$, the model generates the reasoning chain $R := r_1 \dots r_t$ and final answer $y$.
The human raters and judges are now provided with $(p, I, R)$ and asked to rate the visual faithfulness of each $r_i$ in $R$.

\paragraph{Evaluation Task}
Given 300 datapoints $\{(p, I, R)\}_{i=1}^{300}$ a human annotator must rate the visual faithfulness of each $r_i$ in $R$.
Specifically, annotators must check if each $r_i$ is a) a Perception step and b) visually Unfaithful. An example annotation task can be seen in Table~\ref{tab:annotation-ui}.
After completing the task, each annotator produces a list of length 
 containing counts of perception and unfaithful steps per example. These lists are used to compute inter-rater agreement (ICC) between the two human annotators and the VLM judge.

\paragraph{Human and Judge Evaluators}
The annotations were performed by the authors of the paper. As a result, there was no hiring process or demographic screening. All annotators have technical background in vision-language systems and were familiar with the definition of visual faithfulness used in the study.
The VLM judge performs annotations as per Algorithm~\ref{alg:faithfulness-judge}, using the prompt in Table~\ref{tab:judge-prompt}.

\begin{table*}
  \centering
  \begin{tabular}{llcc}
    \toprule
    \multirow{2}{*}{\textbf{Dataset}} & \multirow{2}{*}{\textbf{Prompting Strategy}} & \multicolumn{2}{c}{\textbf{Correlation}} \\ 
    & & \textbf{Perception} & \textbf{Faithfulness} \\
    \midrule
    \multirow{3}{*}{MMMU} & Vanilla & 0.95 & 0.71 \\ 
                          & Grounding & \textbf{0.94} & \textbf{0.7} \\
                          & Grounding + Rationale & 0.94 & \textbf{0.7} \\
    \midrule
    \multirow{3}{*}{MathVista} & Vanilla & \textbf{0.97} & 0.7 \\ 
                               & Grounding & 0.96 & \textbf{0.73} \\
                               & Grounding + Rationale & 0.96 & 0.65 \\
    \midrule
    \multirow{3}{*}{ScienceQA} & Vanilla & 0.79 & 0.48 \\ 
                               & Grounding & \textbf{0.8} & \textbf{0.56} \\
                               & Grounding + Rationale & 0.76 & 0.51 \\
    \bottomrule
  \end{tabular}
  \caption{Assessment of different prompting strategies for \texttt{Claude 4 Sonnet} as a Judge. Grounding the model in the image by prompting it to describe the image results in highest correlation with human judgment.}
  \label{tab:judge-configurations-full}
\end{table*}

\paragraph{Results}
In Table~\ref{tab:judge-comparisons} (Section~\ref{sec:metric}), we show the correlation of various VLM judges with human judgment, aggregated over all 300 datapoints of our evaluation data. In Table~\ref{tab:judge-comparisons-full}, the same results are reported per data split (\texttt{MMMU}, \texttt{ScienceQA}, and \texttt{MathVista}).

On average, reasoning chains contained 5.7 steps, with 3.2 and 3.4 perception steps according to the two human annotators. Both annotators marked an average of 0.6 unfaithful steps. The longest chain contained 17 steps, and the maximum number of unfaithful steps in a single chain was 8.

\paragraph{Testing Various Judge Configurations}
We further test the best judge model (\texttt{Claude 4 Sonnet}) under various prompting strategies. Namely, 
\begin{enumerate}
    \item \textbf{Vanilla}: This is the simplest prompt, which simply describes the annotation task. 
    \item \textbf{Grounding}: This is largely similar to the vanilla prompt, except that the model is first asked to describe the image, before starting the annotation task. This grounds the image, reducing the scope of model hallucinations. The grounding prompt is described in Table~\ref{tab:judge-prompt}.
    \item \textbf{Grounding + Rationale}: In addition to grounding the model in the image, this prompt asks the model to justify each of its labels in the annotation task. 
    \item \textbf{Grounding + Bounding Box Augmentation}: The prompt is augmented with bounding box coordinates of entities in the image. This helps improve the quality of grounding, leading to lesser model hallucinations. We first extract entities from the input prompt using the same model, and then get the coordinates of these entities using the Grounding DINO~\citep{liu2024grounding} object detector.
\end{enumerate}

\begin{figure*}[htbp]
    \centering
    \includegraphics[width=0.49\textwidth]{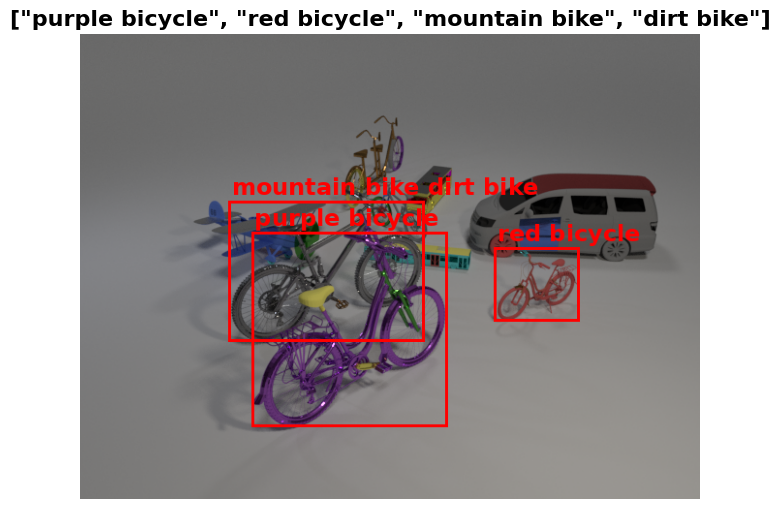}
    \includegraphics[width=0.32\textwidth]{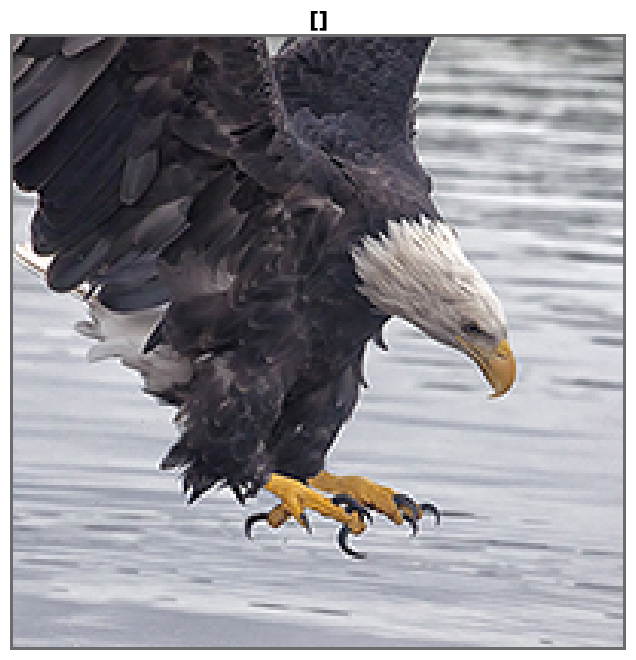}
    \includegraphics[width=\textwidth]{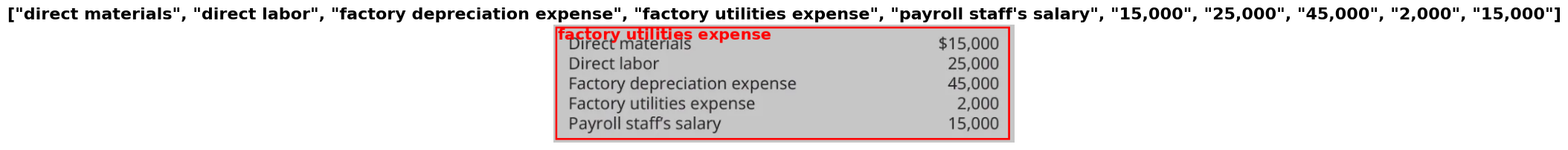}
    \caption{Examples of extracted bounding boxes for VLM judge prompts. Extraction of entities is done through prompting, and corresponding bounding boxes are extracted using Grounding DINO~\cite{liu2024grounding}.
    Extracted entities are listed above each image, while detected objects are marked with red bounding boxes in the image. 
    The pipeline is noisy - both entity extraction and object detection are poor.}
    \label{fig:bb-detector}
\end{figure*}

Using whitespace tokenization (for tokenizer-agnostic comparison), the vanilla prompt averages 256 tokens, the grounding+rationale variant 284 tokens, and the grounding variant 271 tokens.
As seen in Table~\ref{tab:best-judge-configuration}, the Grounding approach has the highest correlation with human judgment. Table~\ref{tab:judge-configurations-full} shows the same result, per split of our evaluation data. We hypothesize that Grounding + Rationale worked poorly since the task became too complex, leading to poorer model attention to the annotation task. Augmentation with bounding boxes was highly noisy, as our entity extraction and object detection modules both introduced noise (Figure~\ref{fig:bb-detector}). Due to the identified bounding box coordinates rarely proving information that would assist the judge model in its task, we remove a formal comparison of this approach with other prompting methods.

\section{More Details on Detection Methods}
\label{sec:appendix-detection-methods}

\begin{figure}[htbp]
    \centering
    \includegraphics[width=0.4\textwidth]{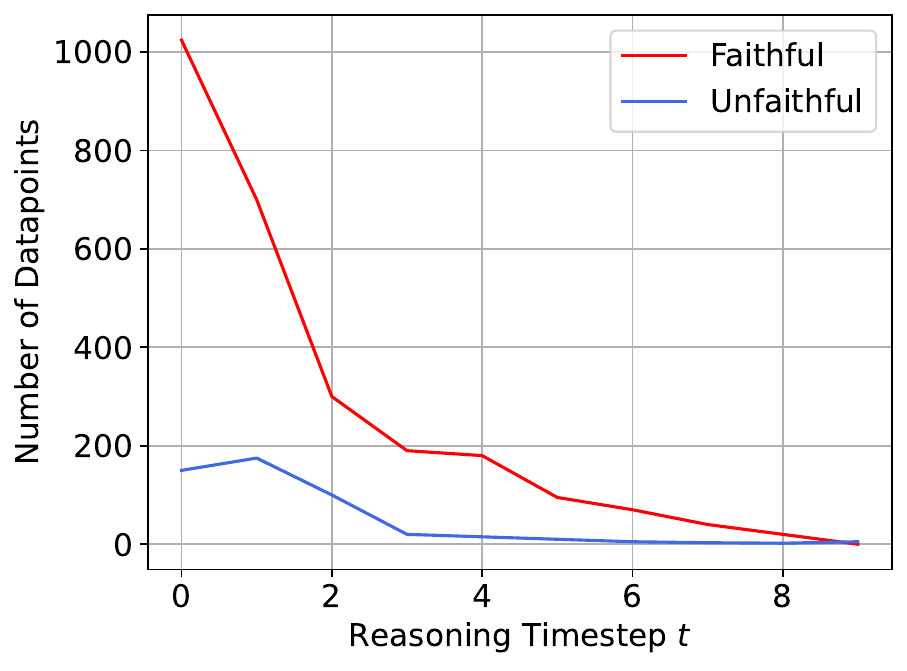}
    \caption{Distribution of \emph{when} dataset.
    The dataset is highly imbalanced - both in the ratio of unfaithful to faithful steps, as well as the number of available steps over time. }
    \label{fig:long-tail-when}
\end{figure}

\begin{figure*}[htbp]
    \centering
    \includegraphics[width=0.9\textwidth]{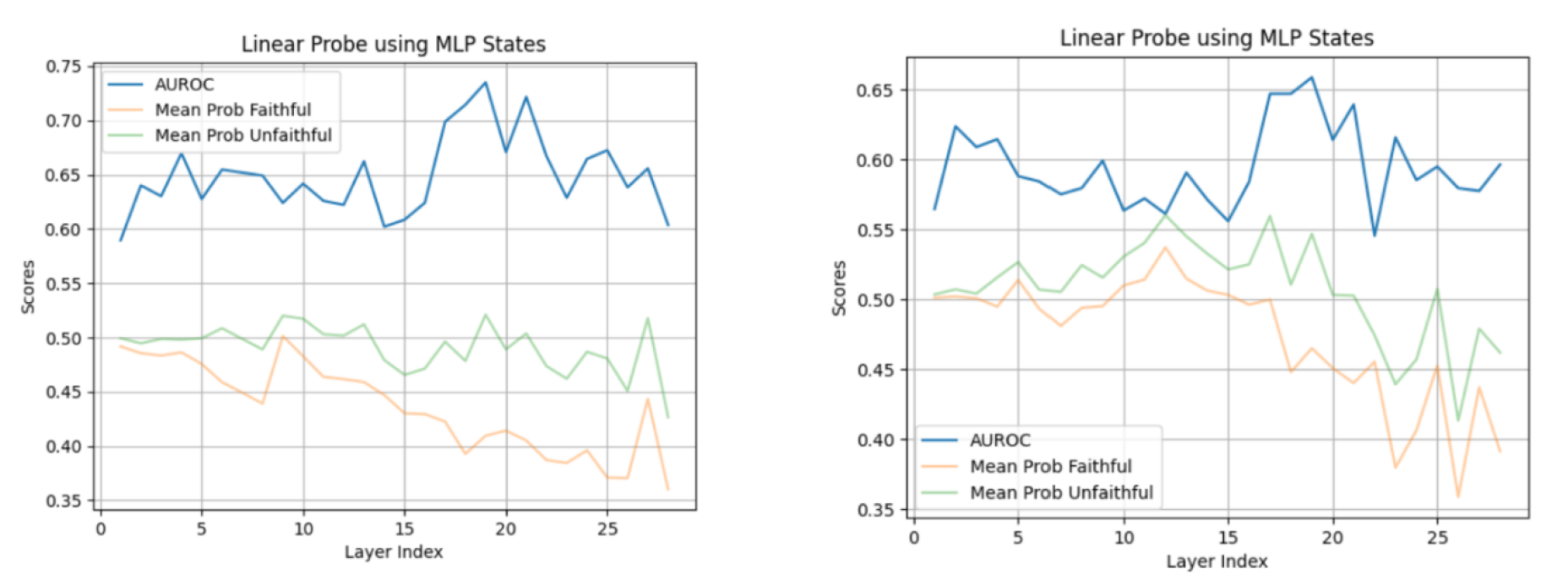}
    \caption{Temporal context drift in hallucination over long reasoning chains.  
    Trained faithfulness detectors may successfully capture the distinction between faithful and unfaithful perception steps early in a model's generation, they fail as the number of steps in the reasoning trace increases. 
    We train a linear probe detector on the embeddings of each layer, on a dataset with short reasoning traces (left) and test it on longer reasoning traces (right). 
    We report the AUROC (y-axis) for each layer's linear probe (x-axis).
    The AUROC drops by around 10 points when evaluated on longer reasoning traces.
    }
    \label{fig:context-drift}
\end{figure*}

\paragraph{Generation of Evaluation Data}
We sample 1200 examples from \texttt{MMEval-Pro}, holding out 500 samples for detector tuning. 
For each sample with prompt $p$ and image $I$, we use a 7B VLM $\theta$ to generate the response -- a reasoning chain $R$ and final answer $y$. 
Following this, we use our metric (as defined in Section~\ref{sec:metric}) to annotate each $r_i$ in $R$ with its type (\texttt{Perception} or \texttt{Reasoning}) and faithfulness (\texttt{Faithful} or \texttt{Unfaithful}). 
Given this labeled dataset, we split it into \texttt{Faithful} and \texttt{Unfaithful} classes. Specifically, for a given reasoning step length $i$, the entire reasoning chain $r_1 \dots r_i$ is classified as faithful or unfaithful depending on the faithfulness of $r_i$.
\begin{align*}
    \mathcal{D_\text{faith}} & \leftarrow (p, I, r_1 \dots r_i^+) \\
    \mathcal{D_\text{unfaith}} & \leftarrow (p, I, r_1 \dots r_i^-) \\
\end{align*}

The dataset is highly imbalanced - both in the ratio of unfaithful to faithful steps, as well as the number of available steps with increasing $i$. This is depicted in Figure~\ref{fig:long-tail-when}. Due to this, we use a small $i$ when creating our dataset ($i\leq2$).

\paragraph{Metrics.} 
We evaluate using F1 against ``gold'' labels produced by \texttt{Claude 4 Sonnet}, applied to generations from the VLM. Using these VLM-judge gold labels, we report F1 on the \emph{unfaithful} class across all detectors.

\paragraph{Detection Methods}
White-box approaches use internal signals such as attention~\citep{zhang2024dhcp,huang2024opera}, logits, or hidden states~\citep{jiang2024interpreting}.
They are training-free but offer coarse, unstable signals due to model complexity~\citep{chen2025survey}.
We use the following white-box approaches: SAPLMA~\citep{azaria2023internal}, HaloScope~\citep{du2024haloscope} and nearest neighbor based detection~\citep{uppaal2023fine}.

Black-box methods rely on surface behavior, including similarity matching, uncertainty~\citep{zhang2024vl},
or trained auxiliary model judgments~\citep{jing2023faithscore,nguyen2025cutpaste,liu2023mitigating,liu2024mmbench,kaul2024throne,wang2023evaluation}.
They generalize better and are more reliable across tasks.
To represent this class of methods, we use simple prompting, as well as leveraging an auxiliary model (\texttt{Claude Sonnet 3.7}).

\paragraph{Degradation of Training Based Detectors due to Temporal Context Drift}
In Section~\ref{sec:when-to-intervene}, we discuss the weaknesses of training based detectors in our setting - while they may successfully capture the distinction between faithful and unfaithful perception steps early in a model's generation, they fail as the number of steps in the reasoning trace increases. This highlights a context drift -- the way the model encodes visual faithfulness changes over time. 
We empirically demonstrate this by training a linear probe detector on a dataset with short reasoning traces ($i\leq2$) and test it on longer reasoning traces ($i>2$). Figure~\ref{fig:context-drift} shows that the AUROC drops by around 10 points when evaluated on longer reasoning traces.
This would not be an issue if data for longer traces were abundant, but as shown in Figure~\ref{fig:long-tail-when}, this distribution is long tailed. 

\section{More Details on the Self-Reflection Method}
\label{sec:appendix-self-reflection-details}

In this section, we include supplementary information to Sections~\ref{sec:mitigation} and~\ref{sec:results}.

\paragraph{Prompts}
The prompt for the detection step of Algorithm~\ref{alg:self-reflection} is described in Table~\ref{tab:prompt-self-reflection-detect} while the regeneration step prompt is in Table~\ref{tab:prompt-self-reflection-regenerate}.

\paragraph{Results on More Models}
In Table~\ref{tab:main-mmvp}, we show consistently strong performance of the self-reflection method, in improving visual faithfulness across various reasoning trained models.

\begin{table}
  \centering
   \resizebox{0.51\textwidth}{!}{
  \begin{tabular}{llcc}
    \toprule
    \textbf{Model} & \textbf{Method} & \textbf{UPR ($\downarrow$)} & \textbf{Acc ($\uparrow$)} \\
    \midrule
    \multirow{2}{*}{\texttt{OpenVLThinker}} & Vanilla & 11.8 & \textbf{50.7} \\ 
                           & + Ours & \textbf{2.3} & 47.7 \\
    \midrule
    \multirow{2}{*}{\texttt{Ocean-R1}} & Vanilla & 8.2 & 41.3 \\ 
                      & + Ours & \textbf{1.4} & \textbf{41.7}\\
    \midrule
    \multirow{2}{*}{\texttt{MM-Eureka}} & Vanilla & 6.9 & 30.7\\ 
                    & + Ours & \textbf{3.1} & \textbf{37.3}\\
    \bottomrule
  \end{tabular}
  }
  \caption{Impact of our method on the visual faithfulness of reasoning chains, on the \texttt{MMVP} dataset. Our method consistently improves UPR, while frequently also improving final answer accuracy.
  All numbers are recorded as percentages.}
  \label{tab:main-mmvp}
\end{table}

\begin{table}
  \centering
   \resizebox{0.51\textwidth}{!}{
  \begin{tabular}{llcc}
    \toprule
    \textbf{Dataset} & \textbf{Method} & \textbf{UPR ($\downarrow$)} & \textbf{Acc ($\uparrow$)} \\
    \midrule
    \multirow{2}{*}{\texttt{MMEvalPro}} & Vanilla & 13.4 & 78.7 \\ 
                       & + Ours & \textbf{4.8} & \textbf{82.8}\\
    \midrule
    \multirow{2}{*}{\texttt{HallusionBench}} & Vanilla & 22.0 & 55.3 \\ 
                            & + Ours & \textbf{7.9} & \textbf{64.3}\\
    \midrule
    \multirow{2}{*}{\texttt{MMVP}} & Vanilla & 8.4 & 35.0\\ 
                  & + Ours & \textbf{1.7} & \textbf{39.7}\\
    \bottomrule
  \end{tabular}
  }
  \caption{Impact of our method on the visual faithfulness of reasoning chains, using the \texttt{ThinkLite-VL} model. Our method consistently improves UPR, while frequently also improving final answer accuracy.
  All numbers are recorded as percentages.}
  \label{tab:main-thinklite}
\end{table}

\begin{table}
  \centering
   \resizebox{0.5\textwidth}{!}{
  \begin{tabular}{lc}
    \toprule
    & \textbf{Number of Reasoning Steps} \\
    \midrule

    Regeneration invoked & 5532 \\  
    Successful regeneration in 1 attempt &  2096 \\
    Successful regeneration in 2 attempts &  794 \\
    Successful regeneration in 3 attempts &  425 \\
    Failure after 3 attempts & 2217 \\
    
    \bottomrule
  \end{tabular}
  }
  \caption{Breakdown of self-reflection outcomes by number of regeneration attempts.
    Most unfaithful steps are corrected within one regeneration, and over 90\% within three.
    The remaining unresolved cases correspond to instances where the model likely lacks the visual knowledge to generate a faithful description, indicating that the reflection process reaches its natural limit rather than incurring inefficiency.}
  \label{tab:mitigation-latency}
\end{table}

\paragraph{Supplementary Tables}
Table~\ref{tab:main-thinklite} (accompanying Figure~\ref{fig:main-result}) shows self-reflection improving reasoning chain visual faithfulness across 
datasets, using the \texttt{ThinkLite-VL} model. 
Table~\ref{tab:mitigation-latency} (accompanying Figure~\ref{fig:self-reflection-latency}) shows the number of unfaithful perception steps, split across the number of regenerations required to correct them.

\begin{table*}[ht]
  \centering    
    \begin{tabular}{lp{10cm}}
    \toprule
    \multicolumn{2}{l}{\textbf{Datapoint provided to Annotator}} \\ 
    \midrule
    \multirow{4}{*}{Prompt $p$} & Which solution has more solutes? \\ 
    & A. Solution A \\
    & B. Solution B \\
    & C. Both have the same amount of solutes \\
    & D. Cannot be determined \\
    \midrule
    
    Image $I$ & \raisebox{-0.5\height}{\includegraphics[width=0.4\textwidth]{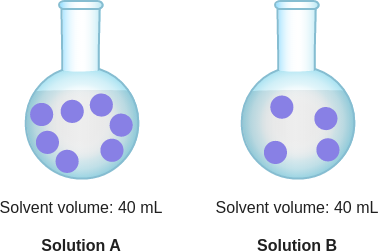}} \\
    \midrule
    
    \multirow{9}{*}{Reasoning Chain $R$} & SENTENCE 1: \\
    & To determine which solution has more solutes, we need to compare the number of solute particles in each solution. \\
    & SENTENCE 2:\\
    & - Solution A has 5 solute particles. \\
    & SENTENCE 3: \\
    & - Solution B has 4 solute particles. \\
    & SENTENCE 4: \\
    & Since Solution A has more solute particles than Solution B, the correct answer is A. \\
    \midrule
    
    \multicolumn{2}{l}{\textbf{Information filled by Annotator}} \\ 
    \midrule
    Number of perception steps	& 2 \\
    Number of unfaithful steps & 1 \\
    
    \bottomrule
    \end{tabular}
    \caption{Example datapoint provided to an annotator for the human correlation study. An annotator (human or VLM judge) is provided with a prompt $p$, image $I$ and reasoning chain $R$, as shown in the example above. Given this, they must simply count the number of Perception and Unfaithful steps that exist in $R$, filling in the final two rows of the table.}
    \label{tab:annotation-ui}
\end{table*}

\begin{table*}
    \centering
    \resizebox{\textwidth}{!}{
    \begin{tabular}{p{15cm}}  
    \texttt{You are an impartial evaluator. Your task is to assess the faithfulness of the reasoning chain provided by a model in response to a visual reasoning task.} \\ \\ 

    \texttt{Each task includes:} \\
    \texttt{- A query, consisting of an image and accompanying text.} \\ 
    \texttt{- A reasoning chain produced by the model (a sequence of sentences)} \\ \\

    \texttt{Before evaluating the reasoning, begin by briefly describing the image.} \\ \\

    \texttt{Your job is NOT to evaluate the final answer, but to analyze each sentence in the reasoning chain.} \\ \\

    \texttt{For each sentence, provide two labels:} \\
    \texttt{1. Type:} \\
    \texttt{- PERCEPTION: The sentence describes or interprets visual content from the image.} \\
    \texttt{- REASONING: The sentence draws logical inferences, conclusions, or relations that go beyond direct visual description.} \\

    \texttt{2. Faithfulness (only if Type = PERCEPTION):} \\
    \texttt{- FAITHFUL: The information accurately reflects what is present in the image.} \\
    \texttt{- UNFAITHFUL: The information misrepresents or contradicts the image.} \\ \\

    \texttt{Instructions:}
    \texttt{- If a sentence is of type REASONING, leave the Faithfulness field blank.} \\
    \texttt{- Focus only on the sentence content; ignore formatting or grammar unless it affects meaning.} \\
    \texttt{- Use the image to verify visual claims.} \\ \\

    \texttt{Your output should be a list of sentences with their corresponding labels, like so:} \\ \\ 

    \texttt{Sentence 1: "<sentence>"} \\
    \texttt{Type: PERCEPTION} \\
    \texttt{Faithfulness: FAITHFUL} \\ \\

    \texttt{Sentence 2: "<sentence>"} \\
    \texttt{Type: REASONING} \\ \\ 

    --- \\ \\

    \texttt{EVALUATION TASK:} \\ \\ 

    \texttt{Query (text only): <query\_text>} \\
    \texttt{[The relevant image is provided separately.]} \\ \\
    
    \texttt{Model's Answer: <model\_answer>} \\ \\ 

    \texttt{Your Verdict:}
    \end{tabular}
    }
    \caption{The prompt template used for evaluating the visual faithfulness in reasoning chains, using a VLM judge model (as described in Algorithm~\ref{alg:faithfulness-judge}).}
    \label{tab:judge-prompt}
\end{table*}

\begin{table*}
    \centering
    \resizebox{0.85\textwidth}{!}{
    \begin{tabular}{p{15cm}}  
    \texttt{You are an impartial evaluator. Your task is to assess the faithfulness of a partial reasoning chain generated by a model in response to a visual reasoning task.} \\ \\

    \texttt{Each task includes:} \\
    \texttt{- A query, consisting of an image and accompanying text.} \\
    \texttt{- A partial reasoning chain, composed of multiple sentences (Sentence 1 to Sentence i).} \\ \\ 

    \texttt{Step 1: Image Description} \\
    \texttt{Begin by briefly describing the image.} \\ \\

    \texttt{Step 2: Sentence Classification} \\ 
    \texttt{Each sentence in the reasoning chain falls into one of two categories:} \\
    \texttt{- PERCEPTION: Describes or interprets visual content from the image.} \\
    \texttt{- REASONING: Draws logical inferences or conclusions beyond direct visual observation.} \\ \\ 

    \texttt{Only PERCEPTION sentences are evaluated for faithfulness:} \\
    \texttt{- FAITHFUL: Accurately reflects the image.} \\
    \texttt{- UNFAITHFUL: Misrepresents or contradicts the image, and the visual detail is relevant or important to the question or reasoning.} \\ \\ 

    \texttt{Evaluation Instructions:} \\
    \texttt{- Assess each sentence in the reasoning chain.} \\
    \texttt{- For PERCEPTION sentences, determine whether they are visually FAITHFUL or UNFAITHFUL.} \\
    \texttt{- A sentence should only be considered UNFAITHFUL if:} \\
    \texttt{        - It misrepresents or contradicts the image, and} \\
    \texttt{        - The visual detail is relevant or important to the question or reasoning.} \\
    \texttt{- Minor or irrelevant visual errors (e.g., small background details, non-essential objects) can be ignored.} \\
    \texttt{- If any PERCEPTION sentence contains a significant unfaithful detail, label the entire chain as UNFAITHFUL.} \\ 
    \texttt{ If multiple sentences are unfaithful, identify and highlight only the first one.} \\
    \texttt{- Assuming the first unfaithful sentence is Sentence k, return:} \\
    \texttt{        - All sentences up to (but not including) Sentence k as the faithful prefix} \\
    \texttt{        - Sentence k as the first unfaithful sentence} \\
    \texttt{- If no sentence is unfaithful:} \\
    \texttt{        - Label the chain FAITHFUL} \\
    \texttt{        - Use the full reasoning chain as the faithful prefix} \\
    \texttt{        - Leave "First unfaithful sentence" blank} \\ \\

    \texttt{OUTPUT FORMAT:} \\ \\
    \texttt{[Faithfulness]: "<FAITHFUL or UNFAITHFUL>"} \\
    \texttt{[Faithful reasoning chain prefix]: "<full prefix or full chain if FAITHFUL>"}  \\
    \texttt{[First unfaithful sentence]: "<first unfaithful sentence or blank if FAITHFUL>"} \\ \\

    \texttt{EVALUATION TASK:} \\ \\ 

    \texttt{Query (text only): <query\_text>} \\
    \texttt{[The relevant image is provided separately]} \\ \\

    \texttt{Reasoning Chain: <partial\_reasoning\_chain>} \\ \\

    \texttt{Your Verdict:}
    \end{tabular}
    }
    \caption{Prompt Template used for the detection step (as described in Algorithm~\ref{alg:self-reflection}).}
    \label{tab:prompt-self-reflection-detect}
\end{table*}

\begin{table*}
    \centering
    \begin{tabular}{p{12cm}}  
    \texttt{You are given a visual question answering task, along with a partially incorrect reasoning chain. The last sentence contains an incorrect description of the image.} \\ \\

    \texttt{Your task is to:} \\
    \texttt{1. Use the image to correct this final sentence.} \\
    \texttt{2. Regenerate only the last sentence, and put it in [ ].} \\ \\ 

    ----- \\ \\ 
    
    \texttt{Example:} \\ \\ 

    \texttt{Question:} \\
    \texttt{Is the woman wearing a hat?} \\
    \texttt{A. Yes} \\
    \texttt{B. No} \\ \\

    \texttt{Partial reasoning chain:} \\
    \texttt{There is a woman in the image. She is standing outside.} \\ \\

    \texttt{Last sentence (with error):} \\
    \texttt{She is wearing a scarf.   ← (This is incorrect)} \\ \\ 

    \texttt{Corrected sentence:} \\
    \texttt{[She is wearing a hat.]} \\ \\ 

    ----- \\ \\ 
    
    \texttt{Instructions:} \\
    \texttt{- Output only the corrected last sentence, and enclose it in brackets.} \\
    \texttt{- Do NOT include any other sentences from the reasoning chain.} \\ \\ 

    \texttt{Now try this one:} \\ \\ 

    \texttt{Question:} \\ 
    \texttt{<query\_text>} \\ \\

    \texttt{Partial reasoning chain:} \\
    \texttt{<faithful\_prefix>} \\ \\ 

    \texttt{Last sentence (with error):} \\
    \texttt{<unfaithful\_sentence>} \\ \\

    \texttt{--- Regenerate the last, corrected sentence below ---}
    
    \end{tabular}
    \caption{Prompt Template used for the Regeneration step by the VLM (as described in Algorithm~\ref{alg:self-reflection}).}
    \label{tab:prompt-self-reflection-regenerate}
\end{table*}

\end{document}